\documentclass[letter,conference]{IEEEtran}
\usepackage{cite}
\usepackage[T1]{fontenc}
\usepackage{amsmath,amssymb,amsfonts}
\usepackage{algorithmic}
\usepackage{graphicx}
\usepackage{textcomp}
\usepackage[binary-units]{siunitx}
\usepackage{xcolor}
\usepackage{nicefrac}
\usepackage{multirow}
\usepackage{subcaption}

\usepackage{url}  
\def\BibTeX{{\rm B\kern-.05em{\sc i\kern-.025em b}\kern-.08em
    T\kern-.1667em\lower.7ex\hbox{E}\kern-.125emX}}
    
\newcolumntype{M}[1]{>{\centering\arraybackslash}m{#1}}
\newcolumntype{R}[1]{>{\raggedleft\arraybackslash}m{#1}}
\newcolumntype{L}[1]{>{\raggedright\arraybackslash}m{#1}}

\begin{document}
\bstctlcite{IEEEexample:BSTcontrol}

\title{LE3D: A Lightweight Ensemble Framework of Data Drift Detectors for Resource-Constrained Devices}

\author{\IEEEauthorblockN{Ioannis Mavromatis\IEEEauthorrefmark{1}, Adrian Sanchez-Mompo\IEEEauthorrefmark{1}, Francesco Raimondo\IEEEauthorrefmark{2}, James Pope\IEEEauthorrefmark{4}, Marcello Bullo\IEEEauthorrefmark{1},\\ Ingram Weeks\IEEEauthorrefmark{1}, Vijay Kumar\IEEEauthorrefmark{1}, Pietro Carnelli\IEEEauthorrefmark{1},  George Oikonomou\IEEEauthorrefmark{2}, Theodoros Spyridopoulos\IEEEauthorrefmark{3}, and Aftab Khan\IEEEauthorrefmark{1}}
\IEEEauthorblockA{\IEEEauthorrefmark{1} Bristol Research and Innovation Laboratory (BRIL), Toshiba Europe Ltd., Bristol, UK\\
\IEEEauthorrefmark{2} Department of Electrical and Electronic Engineering, University of Bristol, Bristol, UK\\
\IEEEauthorrefmark{4} Department of Engineering Mathematics, University of Bristol, Bristol, UK\\
\IEEEauthorrefmark{3} School of Computer Science and Informatics, Cardiff University, Cardiff, UK\\
Emails: \{Ioannis.Mavromatis, Aftab.Khan\}@toshiba-bril.com, \{F.Raimondo, James.Pope\}@bristol.ac.uk}}

\maketitle

\begin{abstract}
Data integrity becomes paramount as the number of Internet of Things (IoT) sensor deployments increases. Sensor data can be altered by benign causes or malicious actions. Mechanisms that detect drifts and irregularities can prevent disruptions and data bias in the state of an IoT application. This paper presents LE3D, an ensemble framework of data drift estimators capable of detecting abnormal sensor behaviours. Working collaboratively with surrounding IoT devices, the type of drift (natural/abnormal) can also be identified and reported to the end-user. The proposed framework is a lightweight and unsupervised implementation able to run on resource-constrained IoT devices. Our framework is also generalisable, adapting to new sensor streams and environments with minimal online reconfiguration. We compare our method against state-of-the-art ensemble data drift detection frameworks, evaluating both the real-world detection accuracy as well as the resource utilisation of the implementation. Experimenting with real-world data and emulated drifts, we show the effectiveness of our method, which achieves up to 97\% of detection accuracy while requiring minimal resources to run.

\end{abstract}


\begin{figure*}[ht]
    \centering
    \includegraphics[width=1\textwidth]{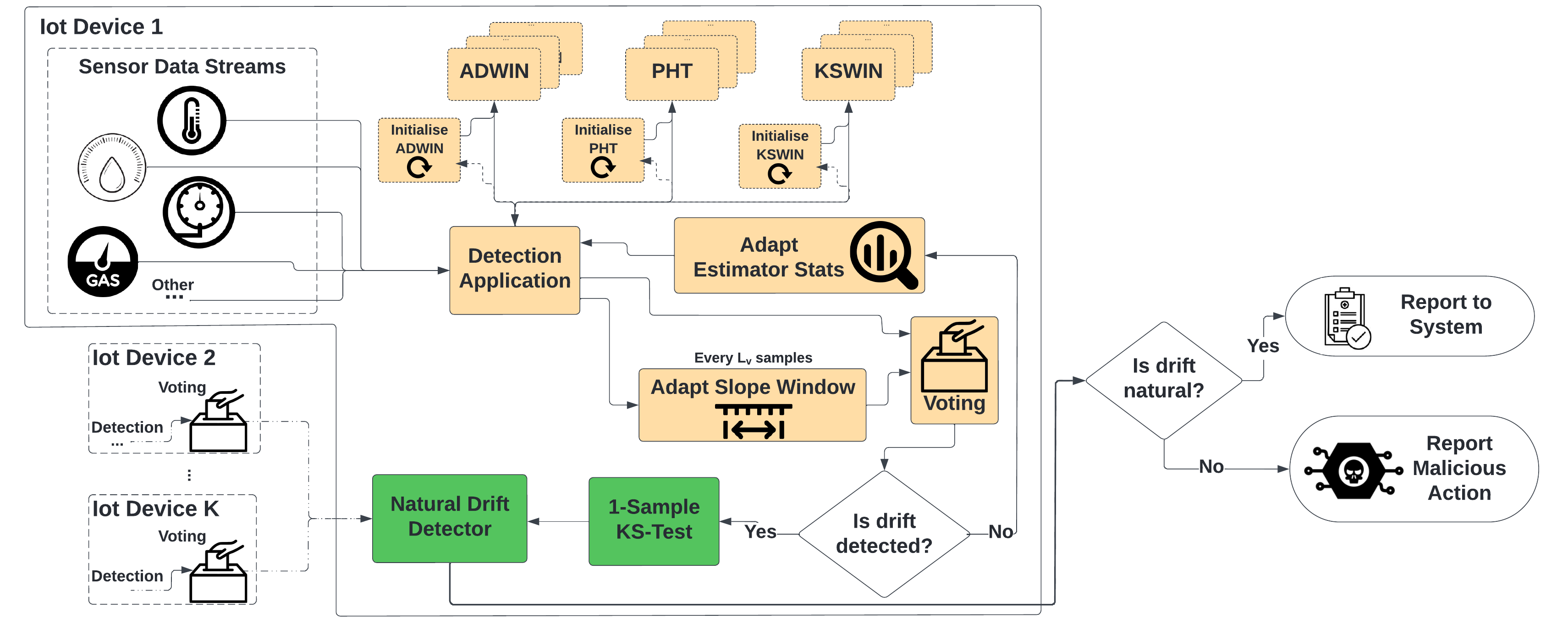}
    \caption{LE3D: the proposed ensemble data drift detection framework. Each IoT device running a detector can process multiple sensor streams and identify abnormalities in the data (functions highlighted in orange). Later, collaboratively with other devices, the type of drift (natural/abnormal) can be classified (functions highlighted in green). The decision can be later reported to the end-user.}
    \label{fig:system_overview}
\end{figure*}

\section{Introduction}\label{sec:intro}
Internet of Things (IoT) is becoming synonymous with everyday computing. It has led to the deployment of billions of interconnected sensors and devices that sense, monitor, and interact with the environments~\cite{iotBillionDevices}. IoT sensors are found in numerous domains, ranging from air pollution monitoring, farming, smart cities, and many more~\cite{IoTbasedSmartCities}. Since these applications rely on the fidelity of the collected data, it is fundamental to preserve the data integrity~\cite{dataIntegrity}. The observed data can be altered by benign causes (e.g., faulty sensors) or malicious actions (e.g., unauthorised data tampering). Both cases can disrupt or bias the states of applications and may result in widespread damage and outages. To prevent that, mechanisms for detecting drifts and irregularities are essential~\cite{mlDataDrift}. Based on this idea, we present Lightweight Ensemble of Data Drift Detectors (LE3D\footnote{Read as ``leed''}), a novel lightweight data drift detection framework able to identify data irregularities in sensor streams. Our framework is publicly available under the repository {\tt\small github.com/toshiba-bril/le3dDataDriftDetector}.

Current IoT systems are characterised by the cost and complexity of their installation, favouring Low-Cost Sensors (LCSs) due to their availability and cost~\cite{sensorComparison}. However, differences between sensor manufacturers, silicon, and installed environments introduce increased variability in the observed data. This leads to inconsistencies when data from different devices are compared~\cite{sensorComparison}. Therefore, as described in~\cite{relativeValuesUsed}, LCSs are better compared using their relative measurements and not their absolute values. Building on this idea, LE3D can detect data stream abnormalities using the observed trends in data streams. Moreover, the comparison against other devices is consolidated upon statistical trends, the goodness of fit, and the relative measurements rather than the absolute values. 


Various Machine Learning (ML)-based drift detection approaches can be found in the literature, e.g.,~\cite{mlDataDrift,mlAutoencoder}. In such approaches, models ``learn'' about the abnormal behaviour from a given dataset, or train on normal data and classify everything else as abnormal. However, when new variables are introduced, ML models tend to require retraining to improve their accuracy~\cite{onlineTraining}. Furthermore, this incremental online learning is usually not possible on resource-constrained IoT devices. Instead, the data are sent to a more powerful device (e.g., a cloud server), the model is updated and is returned to the device for inference. However, this not only increases the communication cost due to the increased data exchange but can introduce additional threats during the transit (e.g., data tampered with, lost, leaked, etc.)~\cite{dataPrivacy}. To overcome these limitations, LE3D provides an online training mechanism that runs on-the-fly and directly on the resource-confined IoT device while requiring minimal resources.


All of the above pave the way for this paper's contribution. LE3D is a framework able to identify irregularities in sensor streams. Operating as an ensemble framework, decisions are based on three estimators, these being the ADaptive WINdowing (ADWIN)~\cite{adwin}, Page-Hinkley Test (PHT)~\cite{pageHinkley}, and Kolmogorov-Smirnov Windowing (KSWIN)~\cite{kswin} algorithms. Even though these estimators individually do not achieve ideal performance, an adaptive voting mechanism leveraging their decisions enhances the framework's accuracy. Our framework is built and optimised for real-world resource-confined IoT devices, introducing minimal overheads and dynamically adapting to new sensor types and streams. Working as a distributed system later, detected drifts can be further classified as natural (e.g., a temperature is increased between day and night) or abnormal (e.g., only a single sensor reporting different than expected temperature).



The rest of the paper is structured as follows. Sec.~\ref{sec:related_work} summarises various solutions found in the literature and describes how LE3D extends the state-of-the-art. Sec.~\ref{sec:framework} describes our data drift detection framework, the drift estimation mechanisms and their configuration steps. Then, our real-world implementation is briefly described in Sec.~\ref{sec:realworld}, where the system architecture and sensor data collected are presented. Our performance investigation can be found in Sec.~\ref{sec:results}. Finally, our work concludes in Sec.~\ref{sec:conclusion} with our final remarks.

\section{Related Work}\label{sec:related_work}
Related frameworks are found in the literature. An optimised ML approach to detecting botnets has been presented in~\cite{botnetsAttack}. Based on decision trees and Bayesian optimisation with Gaussian Process algorithms, this work achieves an accuracy of $>99\%$. In~\cite{offlineClassifiers}, authors present an ensemble framework based on offline classifiers and imbalanced data that achieved $94\%-97\%$ accuracy for the different data classes. Even though both~\cite{botnetsAttack} and~\cite{offlineClassifiers} achieve high accuracy, they are based on offline learning approaches not adaptable to the fast-changing environments of an IoT ecosystem. LE3D is an online learning framework able to adapt and accommodate new data streams fed into the system in real-time. 

The online drift detection approach in~\cite{mlDataDrift} uses deep learning techniques and achieves an accuracy of \textasciitilde$96\%$. However, several thousands of training samples are required for training, as well as increased training time. Our approach relies on just a few tens of samples for the detectors' initialisation without compromising the accuracy. A lightweight Performance Weighted Probability Averaging Ensemble (PWPAE) framework is presented in~\cite{PWPAE}. PWPAE is a four-party supervised ensemble data drift detector backed by adaptive weights that change in real-time. LE3D moves a step further and, using the individual decisions of each detector, can collaboratively later classify the drift as normal or abnormal. Moreover, being an unsupervised method makes it optimal for resource-constrained deployments. Due to the similarities between our framework and PWPAE, we will use it for our performance comparison.

\section{Proposed Framework}\label{sec:framework}
Consider an indoor air quality monitoring use case as an example IoT application. Example of sensor data collected are the environmental temperature, humidity, pressure, Volatile Organic Compounds (VOC), etc. All these readings can vary greatly in terms of their absolute values and standard deviation. For example, the average temperature for an airconditioned room could be between \SIrange[range-phrase=--,range-units=single]{20}{22}{\celsius} with a standard deviation of \SIrange[range-phrase=--,range-units=single]{1}{2}{\celsius}, the average humidity could be between \SIrange[range-phrase=--]{40}{50}{\%} with a standard deviation of \SIrange[range-phrase=--]{4}{6}{\%}, and the pressure can vary between \SIrange[range-phrase=--,range-units=single]{99}{103}{\kilo\pascal} with a standard deviation of \SIrange[range-phrase=--,range-units=single]{200}{250}{\pascal}~\cite{indoorairpollution}. When considering the data distributions of the above data, it is evident that they are not easily generalisable. 

\subsection{System Overview}
LE3D framework works as a two-layer hierarchical system. Fig.~\ref{fig:system_overview} provides an overview of the proposed framework. Initially, each IoT device is responsible for detecting drift in individual sensor streams; at this stage, the edge device cannot distinguish between natural and malicious drifts. When a new sensor stream is detected, all the estimators' statistical hyperparameters are fine-tuned using a two-step grid search. Later, for each sample received the three estimators (ADWIN, PHT, and KSWIN) individually detect whether a drift has occurred. Using an adaptive sliding window and the decisions from all estimators, a detector decides whether the drift is valid or not (voting with equal weights for each estimator). The adaptive window and the voting can enhance the effectiveness and efficiency of the individual estimators.

As a second step, collaboratively with its surrounding neighbouring IoT devices, an IoT node can further classify a drift as natural (i.e., when a similar drift is observed in a number of devices with similar properties) or abnormal (i.e., when only a single sensor stream on a single device presents drifting behaviour) based on the outcome of voting decisions and the statistical significance of the drift. More detailed definitions of natural and abnormal drifts are given in Sec.~\ref{subsec:naturalmalicious}. This decision is later reported to a backend system for inspection and mitigation by an end-user. In the following sections, we describe in more detail the individual system components and their functionality.

\subsection{Different Types of Data Drift}
Data drift when the data distribution changes in a non-stationary environment. Table~\ref{tab:key_notation} summarises the key notation used in the paper for easier comprehension. There exists an index $N \in \mathbb{N}^*$ of the sequence $\mathcal{X}$ such that all samples $x_i \in \mathcal{X}_{1:N-1} = \left \{ x_i \right \}_{i=1}^{N-1}$ share the same stability properties (e.g., same probability distribution). Sequence $\mathcal{X}$ denotes a single sensor stream arriving at an IoT device. A sample with index $N$ is the sample where a sudden or continuous drift occurs.  Values can stabilise again after a number of samples $k \in \mathbb{N}^*$  and converge to the same or a new stability concept. The instances between $\mathcal{X}_N$ and $\mathcal{X}_{N+k}$ are considered to be drifting. According to the length of $k$, different types of drift can be described. 

\begin{figure}[t]
    \centering
    \includegraphics[width=1\columnwidth]{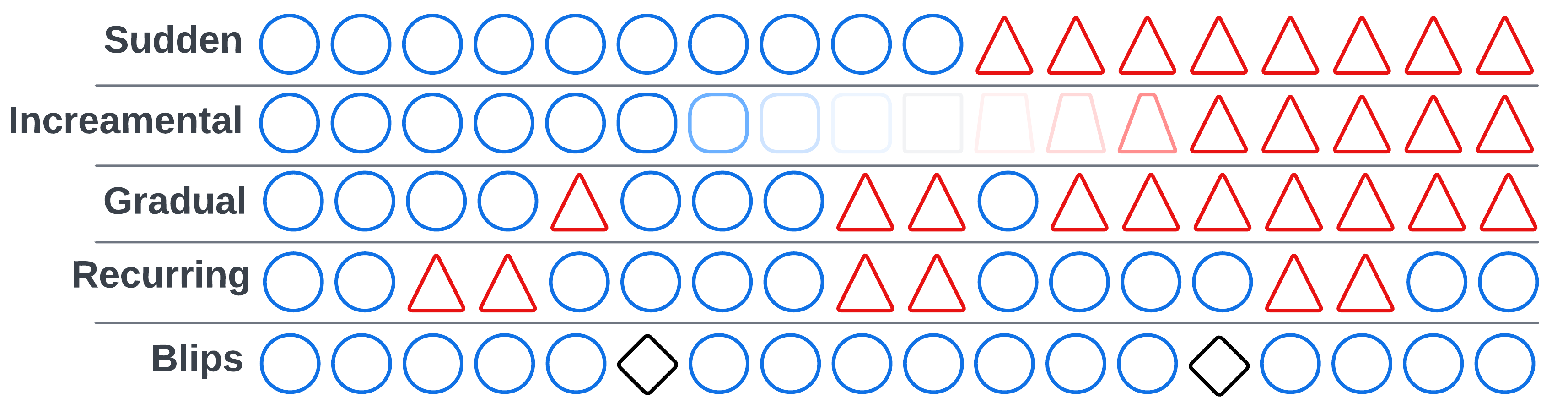}
    \caption{An example of all potential drift types.}
    \label{fig:drifttypes}
\end{figure}

When the samples before and after the drift stabilise to two different concepts and the drift occurs suddenly, i.e., $k = 1$, the drift is considered abrupt~\cite{typesOfDrift}. Gradual and incremental drifts appear when the changes occur steadily. An incremental drift happens when the observed values change progressively between $\mathcal{X}_N$ and $\mathcal{X}_{N+k}$, moving from one stability concept to another. Gradual is the drift where a new concept gradually replaces an old one after $k$ samples. Values in gradual drift alternate between two or more stability concepts and stabilise to one. When the instances of a stability concept appear for a short period and disappear afterwards, the drift is considered recurring. Finally, a blip drift event appears when a single sample is outside the stability concept (i.e., all samples before and after that follow the same distribution) and is considered an outlier. Examples of the drift types can be seen in Fig.~\ref{fig:drifttypes}. More information about the types of drift can be found in~\cite{typesOfDrift}.

Our performance evaluation is based on two types of drift, i.e., abrupt and incremental. These are the most prevalent types found in time-series sensor data. Recurring drifts are a priory considered and are reported as independent events. Gradual drift detection is more suitable for categorical data, while incremental drift suits more time-series data. For this paper, we focus on raw time-series data. Support for categorical data drift is considered a future extension to LE3D. Finally, blip drift events are considered outliers and are ignored from our framework.

\begin{table}[t] 
\renewcommand{\arraystretch}{1.12}
\centering
    \caption{Key Notations.}
    \vspace{1mm}
    \begin{tabular}{R{1.4cm}|L{6.5cm}}
    \textbf{Notation} & \textbf{Explanation}  \\ \hline \hline
    $\mathcal{X}$ & Sequence of sensor samples. \\ 
    $N$ & Index of last non-drifted sample. \\ 
    $k$ & Number of sensor samples drifting. \\ 
    $x_i$ & Sensor sample with index $i$. \\ 
    $W_{\mathbb{A}}$, $W_{\mathbb{R}}$ & Sliding windows (ADWIN, KSWIN). \\
    $L_{\mathbb{A}}$, $L_{\mathbb{R}}$ & Length of sliding windows $W_{\mathbb{A}}$, $W_{\mathbb{R}}$. \\
    $W_{\mathrm{v}}$ & Adaptive sliding window for voting. \\
    $L_{\mathrm{v}}$ & Length of voting window $W_{\mathrm{v}}$. \\
    $W_{\mathrm{m}}$ & Non-overlapping sliding window for trend calculations. \\
    $L_{\mathrm{m}}$ & Length of trend calculation window $W_{\mathrm{m}}$. \\
    $\mu_{\mathbb{A}}$, $\mu_{\mathbb{P}}$, $\mu_{\mathbb{K}}$ & Mean sample value of different estimators. \\
    $\sigma_{\mathbb{A}}$, $\sigma_{\mathbb{P}}$, $\sigma_{\mathbb{K}}$ & Std. Deviation of sample value of different estimators. \\
    $\mu_{W_{\mathrm{m}}}$ & Mean value of samples in window $W_{\mathrm{m}}$. \\
    $\theta_{W_{\mathrm{m}}}$ & Mean value of samples in window $W_{\mathrm{m}}$. \\
    $b$ & Number of samples used for initialising an estimator. \\
    $\mathcal{E}$ & List of estimators for sequence $\mathcal{X}$. \\
    $\mathcal{V}$ & Voting decisions of all estimators and collectively. \\
    $\mathcal{D}$ & List of detectors in the system. \\
    $\mathcal{S}$ & List of sensor streams fed into each detector. \\
    \end{tabular}
\label{tab:key_notation}
\end{table}

\subsection{Different Detection Algorithms}\label{subsec:detectors}
Various lightweight drift estimators are presented in the literature. Some rely on continuous data stream (e.g., a  time-series of temperature data), e.g., ADWIN~\cite{adwin}, while others, e.g., Hoeffding Drift Detection Method (HDDM)~\cite{hddm_w}, work with discrete values and real predictions. The nature of the sensor types considered in our system and the absence of knowing whether there is a drift led us to consider estimators from the first category. The three detection algorithms used are ADWIN~\cite{adwin}, PHT~\cite{pageHinkley}, and KSWIN~\cite{kswin} (based on the Kolmogorov-Smirnov (KS) statistical test). In the future, if more estimators are required for different use-cases can be easily integrated into LE3D.

\subsubsection{ADaptive WINdowing (ADWIN) algorithm}
ADWIN can detect distribution changes and drifts in data that vary with time. It uses an adaptive sliding window $W_{\mathbb{A}}(n,L_{\mathbb{A}}) = \mathcal{X}_{n:n+L_{\mathbb{A}}}$ with $n,L_{\mathbb{A}} \in \mathbb{N}^*$, that is recalculated online according to the rate of change observed from the data $\mathcal{X}_{n:n+L_{\mathbb{A}}} \subset \mathcal{X}$. $W_{\mathbb{A}}$ is discretised in two sub-windows $W_{\mathbb{A}} = [W_{\mathrm{hist}},W_{\mathrm{new}}]$ with $W_{\mathrm{hist}}(n,L_{\mathrm{hist}}) = \mathcal{X}_{n:n+L_{\mathrm{hist}}}$ and $W_{\mathrm{new}}(L_{\mathrm{hist}},L_{\mathrm{new}}) = \mathcal{X}_{L_{\mathrm{hist}}:L_{\mathrm{hist}}+L_{\mathrm{new}}}$ such that $L_{\mathrm{hist}} + L_{\mathrm{new}} = L_{\mathbb{A}}$. When a new sample is received, ADWIN examines all possible cuts for $W_{\mathbb{A}}$ calculating the mean values $\mu_{\mathrm{hist}}$ and $\mu_{\mathrm{new}}$ and the absolute difference $\phi_{\mathbb{A}} = |\mu_{\mathrm{hist}} - \mu_{\mathrm{new}}|$. The optimal lengths $L_{\mathrm{hist}}$ and $L_{\mathrm{new}}$ for the two sub-windows are given comparing a threshold $\epsilon_{\mathrm{cut}}$ against all values $\phi_{\mathbb{A}}$. It is given from $\epsilon_{\mathrm{cut}} > |\max(\phi_{\mathbb{A}})|$. Finally, $\epsilon_{\mathrm{cut}}$ is defined as:
\begin{subequations}\label{eq:problem}
\begin{align}
\displaystyle m = \frac{1}{1/L_{\mathrm{hist}}  + 1/L_{\mathrm{new}}}, \quad&\textrm{and}\quad \delta^{\prime} = \frac{\delta}{W_{\mathbb{A}}}\\
\epsilon_{\mathrm{cut}} = \sqrt{\frac{2}{m}  \: \sigma^2_{W} \: \frac{2}{\delta^{\prime}}} &+ \frac{2}{3m}\ln{\frac{2}{\delta^{\prime}}} &
\end{align}
\end{subequations}
where $\sigma^2_{W}$ is the observed variance of the elements in window $W_{\mathbb{A}}$ and $\delta \in (0,1)$ is the user-defined confidence value. Once a drift is detected, all the old data samples within $W_{\mathrm{hist}}$ are discarded. ADWIN can effectively detect gradual drift since the sliding window can be extended to a large-sized window and identify long-term changes. Abrupt changes can again be identified with a small number of samples due to the big difference introduced in the mean values. More information about ADWIN can be found in~\cite{adwin}.

\subsubsection{Page-Hinkley Test (PHT) algorithm}
PHT is a variant of the CUmulative SUM (CUSUM) test. It has optimal properties in detecting changes in the mean value of a normal process. By default, PHT is a one-sided drift detector and only detects changes when the mean increases. We extended PHT with symmetry to work as a two-sided estimator for our implementation. For every sample received, PHT recalculates the mean value $\mu_{\mathbb{P}}$ and the cumulative sum: 
\begin{align}
U(i) = \sum_{i = 0}^{N} \left(x_i - \mu_{\mathbb{P}}^{i} - \frac{\beta}{2} \right)
\end{align}
where $\beta \in \mathbb{R}^{+}$ is user-defined and $\mu_{\mathbb{P}}^{0} = 0$. 

The result of the estimator is given by $\max(U) - U(i) \geq \lambda$, indicating an increase in the observed mean value, or by $U(i) - \min(U) \geq \lambda$, indicating a decrease in $\mu_{\mathbb{P}}^{i}$. $\lambda \in \mathbb{N}^*$ is a user-defined threshold. The magnitude of $\beta$ describes the tolerated changes that will not raise an alarm, while $\lambda$ tunes the false alarm rate. Larger $\lambda$ entails fewer false-positives detections while increasing the false negatives. PHT easily identifies abrupt drifts due to the sudden change in the mean value. In contrast, incremental drift can be identified by sporadically sampling the time-series stream data.

\subsubsection{Kolmogorov-Smirnov Windowing (KSWIN) algorithm}
Our final estimator is KSWIN and is based on a KS statistical test. KS test is a non-parametric test, accepting one-dimensional data and operating with no assumption of the underlying data distribution. KSWIN maintains a fixed size sliding window $W_{\mathbb{K}}(n,L_{\mathbb{K}}) = \mathcal{X}_{n:n+L_{\mathbb{K}}}$ with $n,L_{\mathbb{K}} \in \mathbb{N}^*$. $W_{\mathbb{K}}$ is discretised in two sub-windows $W_{\mathbb{K}} = [W_{\mathrm{\Omega}},W_{\mathrm{R}}]$ with $W_{\mathrm{\Omega}}(n,L_{\mathrm{\Omega}}) = \mathcal{X}_{n:n+L_{\mathrm{\Omega}}}$ and $W_{\mathrm{R}}(L_{\mathrm{\Omega}},L_{\mathrm{R}}) = \mathcal{X}_{L_{\mathrm{\Omega}}:L_{\mathrm{\Omega}}+L_{\mathrm{R}}}$ such that $L_{\mathrm{\Omega}} + L_{\mathrm{R}} = L_{\mathbb{K}}$.  A two-sampled KS test is performed on $W_{\mathrm{R}}$ and $W_{\mathrm{\Omega}}$. It compares the absolute distance $dist_{W_{\mathrm{\Omega}},W_{\mathrm{R}}}$ between two empirical cumulative data distribution, i.e., $dist_{W_{\mathrm{\Omega}},W_{\mathrm{R}}} = \sup\limits_x|F_{W_{\mathrm{R}}}(x) - F_{W_{\mathrm{\Omega}}}(x)|$ where $\sup\limits_x$ is the least upper bound of the distance. $F_{(\cdot)}(x)$ represents the empirical distribution function.

The result of the estimator is given from $dist_{W_{\mathrm{\Omega}},W_{\mathrm{R}}} > \sqrt{-\nicefrac{\ln\alpha}{L_{\mathrm{R}}}}$, where $\alpha \in (0,1)$ defines the parameter sensitivity of the test statistic and is user-defined. Data with increased periodicity and a large window make KSWIN too sensitive and return many false positives. Relatively small $L_{\mathrm{R}}$, i.e., $L_{\mathrm{R}} \approxeq 30$, and an optimised $\alpha$ significantly improve the performance. As described in~\cite{kswin}, KSWIN is capable of detecting gradual and abrupt drifts but falsely classifies many samples as false positives. However, considering the criticality of an error, false positives are not as critical and can be removed in post-processing.

\subsection{Voting Mechanism for Enhanced Detection Performance}\label{subsec:voting}
Let $\mathcal{E} \triangleq \left \{1,\ldots,E \right \}$ with $E \in \mathbb{N}^*$ define the estimators for a sequence $\mathcal{X}$. Considering the estimators from Sec.~\ref{subsec:detectors}, we have $\mathcal{E} \triangleq \left \{1,2,3 \right \}$ in our system. A single estimator is not always able to accurately detect all drifts. We enhance the performance of the framework by introducing a more systemic approach. More specifically, for each $x_i$, our framework updates the statistics of all $\mathcal{E}$. Each $\mathcal{E}$ decides whether $x_i$ is normal or abnormal. Operating on a per-sample fashion, it is unlikely that a single sample will be flagged as abnormal by more than one $\mathcal{E}$. We solve this problem by introducing an adaptive sliding window $W_{\mathrm{v}}(n,L_{\mathrm{v}}) = \mathcal{X}_{n:n+L_{\mathrm{v}}}$ with $n,L_{\mathrm{v}} \in \mathbb{N}^*$, $L_{\mathrm{v}}>L_{\mathbb{K}}$ and a voting mechanism within this window. Doing so, we holistically examine the behaviour of the last $L_{\mathrm{v}}$ samples and can decide whether a drift occurred or not. For all $\mathcal{E}$ we maintain a sequence $v_i \in \mathcal{V}_{1:N}^{\mathcal{E}} = \left \{ v_i \right \}_{i=1}^{N}$ with $N \in \mathbb{N}^*$, $N \geq L_{\mathrm{v}}$, and $v_i \in \{0,1\}$, where $0$ demonstrates normal behaviour and $1$ an abnormal sample. 

From all $\mathcal{E}$ we collectively decide whether the last $L_{\mathrm{v}}$ samples present a drift. All $\mathcal{E}$ participate in the voting with equal weights and with the condition that:
\begin{align}
\mathcal{V}_{W_{\mathrm{v}}} = 
  \begin{cases}
      1,~\mathrm{if}~ \sum\limits_{i \in N} \mathcal{V}^e(v_i) \geq 2,~\forall e \in \mathcal{E} \\
     0,~\mathrm{otherwise}
  \end{cases}
\end{align}
If the majority of $\mathcal{E}$ reported a drift, the values within $W_{\mathrm{v}}$ are perceived as drifted.

\subsection{Adaptive Voting Window Length}

$W_{\mathrm{v}}$ is adaptively modified according to the mean value of the data received. This ensures that all types of drifts will be correctly identified. To adapt $L_{\mathrm{v}}$, we utilise a separate window of values, where the trend of the data is calculated. More specifically, we define a non-overlapping sliding window $W_{\mathrm{m}}(n,L_{\mathrm{m}}) = \mathcal{X}_{n:n+L_{\mathrm{m}}}$ with $n,L_{\mathrm{m}} \in \mathbb{N}^*$, grouping the last $L_{\mathrm{m}}$ samples. Applying a linear least-squares regression, we calculate the slope $s$ with the best goodness of fit. We later calculate the trend of the data $\theta_{W_{\mathrm{m}}} = \arctan{s}$ (measured in degrees $^{\circ}$). When $\theta_{W_{\mathrm{m}}} < 0$, the data trend is downwards; thus, their mean $\mu_{{W}_{\mathrm{s}}}$ value is expected to decrease. Similarly, $\theta_{W_{\mathrm{m}}} > 0$ implies an increase in $\mu_{{W}_{\mathrm{s}}}$. A small drift of any type will introduce a small change in the mean $\mu_{{W}_{\mathrm{s}}}$. These drifts are more difficult to be identified and require a larger $W_{\mathrm{m}}$ to observe a drift. On the other hand, a sharp change is easily detected and is usually associated with an abrupt drift. Thus a smaller $W_{\mathrm{m}}$ can be used for that. Finally, using $\Upsilon = \nicefrac{|\mu_{{W}_{\mathrm{s}}} - \mu_{{W}_{\mathrm{s}}}^\prime|}{\mu_{{W}_{\mathrm{s}}}^\prime}$ we correlate $\mu_{{W}_{\mathrm{s}}}$ with the mean value of the previous window $\mu_{{W}_{\mathrm{s}}}^\prime$.

For all sensor types in our system, we ran a non-linear regression analysis to model the relationship between the window size $W_{\mathrm{v}}$ and $\Upsilon$. The exponential equation:
\begin{align}
    \Upsilon(x) = \zeta \exp(\eta \, x) + \gamma
\end{align}
with coefficients $\zeta$, $\eta$ and $\gamma$, was chosen after the nonlinear regression. This equation achieves high Root Mean Square Error (RMSE) and r-squared $r^2$ for all the different sensor streams; thus, it was considered the best fit for our system. In Table~\ref{tab:nonLinearRegression}, we present the statistical test measures for the above equation and the coefficients found for each sensor stream used in our experimentation. The dataset used for the analysis is described in Sec.~\ref{subsec:uob_sec}.

\begin{table}[t] 
\renewcommand{\arraystretch}{1.12}
\centering
    \caption{Non-Linear Regression and Coefficients.}
    \vspace{1mm}
    \begin{tabular}{r||c|c|c||c|c|c}
    \textbf{} & \multicolumn{3}{c||}{\textbf{Measure Tests}} & \multicolumn{3}{c}{\textbf{Coefficients}}  \\ 
    
    \textbf{Sensor} & \textit{RMSE} & $\chi^2$  & $r^2$ & $\zeta$ & $\eta$ & $\gamma$ \\ \hline \hline
    Temperature & $\scriptstyle42.599$  & $\scriptstyle52626.14$ & $\scriptstyle0.923$ & $\scriptstyle8.782$ & $\scriptstyle-5.021$ & $\scriptstyle1.468$ \\
    Humidity & $\scriptstyle34.226$ & $\scriptstyle33972.54$ & $\scriptstyle0.954$  & $\scriptstyle9.641$ & $\scriptstyle-4.117$ & $\scriptstyle1.508$ \\
    Pressure & $\scriptstyle28.065$ & $\scriptstyle22842.04$ & $\scriptstyle0.961$ & $\scriptstyle7.590$ & $\scriptstyle-5.132$ & $\scriptstyle1.829$
    \end{tabular}
\label{tab:nonLinearRegression}
\end{table}

\subsection{Natural and Abnormal/Malicious Drift}\label{subsec:naturalmalicious}
Natural is considered a drift observed in several devices sharing a common sensor type (e.g., temperature sensor) and characteristics (e.g., installed in the same room). On the other hand, abnormal is the drift detected only on a single sensor stream reported from a single IoT device. An abnormal drift can be considered malicious if triggered by a malevolent action. By sharing only the outcome of the voting mechanism, we can ensure data confidentiality and integrity (as the data never leave the device), cross-validate the results in a distributed fashion and classify the type of drift observed. 

Let $\mathcal{D} \triangleq \left \{1,\ldots,D \right \}$ with $D \in \mathbb{N}^*$ define the detectors in our system. Let $\mathcal{S} \triangleq \left \{1,\ldots,S \right \}$ with $S \in \mathbb{N}^*$ define the number of sensor streams per device. In our architecture, we assume that each IoT device runs a single $\mathcal{D}$ for each sensor stream $\mathcal{S}$ fed to the device. As discussed in Sec.~\ref{subsec:voting}, for each $\mathcal{S}$ there are three estimators that calculate $\mathcal{V}_{W_{\mathrm{v}}}$. Using the outcome of the estimators and the voting mechanism, a detector calculates a one-sample KS test using $Z(x) = x_{\mathrm{ks}} - \mu^\prime$ where $x_{\mathrm{ks}} \in W_{\mathrm{v}}$ and $\mu^\prime$ is the mean value used for initialising the estimator.

Each detector later shares $\mathcal{V}_{W_{\mathrm{v}}}$, $Z$, the sensor type, and some pre-defined metadata (e.g., room number, the zone of the building a sensor is installed, the sensor model, etc.) with neighbouring nodes. The discovery of the neighbouring nodes is out of the scope of this work. Traditional routing protocols and Data Distribution Service (DDS) buses can provide such functionality. Using this information, devices can later decide whether the perceived drift is natural or not. 

The cross-validation of the observed drift relies on the metadata exchanged, the voting decision outcome, and the result of the one-sample KS test. With regards to the metadata, sensor streams with common properties (i.e., the same metadata reported) are expected to be compared. Regarding the KS test, if the distance observed is statistically insignificant for all the received streams, we assume the drift is natural. On the other hand, if the outcome of a specific KS test presents a statistical significance compared to the rest, this sensor drifts abnormally. The detectors then report the observed behaviour to a backend system for further investigation by a system administrator. The cross-correlation of the metadata between different sensors is outside of the scope of this work.

\subsection{Estimator Initialisation}\label{subsec:initialisation}
As discussed in Sec.~\ref{subsec:detectors}, each estimator requires a number of input hyperparameters during its initialisation. Later, all estimators update their statistical models based on the received samples. For LE3D, we introduce an initialisation phase where the ``normal'' behaviour is established for each estimator.

This is done in two different ways. We can either consider the first $b$ received samples (assuming that no drift occurs during this course) or by using a ``trusted'' dataset that accompanies the detector. As described by the Central Limit Theorem (CLT), the distribution of the sample means approximate a normal distribution as the sample size gets larger, regardless of the population's distribution. Based on CLT, and as described in~\cite{samplingSize}, this sample size can be between \SIrange{30}{50}{}. Our system considers the first $100$ samples for the initialisation. Based on them, we calculate the $\mu^\prime$ and the variance $\sigma^2$ and run an exploratory investigation to find the best initialisation parameters for each estimator. 

Our exploration is based on two grid searches. The first one narrows the search space returning an estimated value for the hyperparameters. The second micro-grid search fine-tunes all hyperparameters by evaluating various parameters within a smaller, more precise exploration space. Given the simplicity of the estimators, the notion of ``error score'' is not introduced in our optimisation. Instead, a pre-defined logic is hardcoded in the system. For example, higher $\delta$ values for ADWIN increase the sensitivity of the estimator. Our chosen $\delta$ is a value that does not return candidate drift within a ``normal'' sample distribution. Similarly, the lowest values for $\lambda$ and $\beta$ are preferred for PHT. For KSWIN, lower $\alpha$ improves the estimator's confidence, while a value of $L_{\mathrm{\Omega}} \approxeq 30$ is preferred for the statistical window. Fixing  $L_{\mathrm{\Omega}} = 30$ we later fine-tune $L_{\mathrm{R}}$ and $\alpha$ as before. Based on the above, our estimators become sensitive enough to detect any introduced drift. Finally, the precision of the grid search (defining the grid steps) can be fine-tuned based on the available hardware and the time constraints of each use case.

\section{LE3D: Framework Implementation}\label{sec:realworld}

\subsection{System Architecture and Implementation}\label{subsec:implementation}

We assume a standard three-tier architecture: cloud, edge and endpoint. At the top, a central ``cloud server'' component is responsible for the application and service deployment to the edge/endpoint tiers and for visualising the results. The ``edge'' tier consists of several resource-constrained IoT devices, acting as the ``edge'' nodes deployed close to the endpoints or sensors. The ``edge'' tier accommodates the data bus and message protocols, the detection and voting mechanisms and is responsible for sending the natural or abnormal drift decisions to the cloud. Finally, our endpoints collect and disseminate the sensor data to the edge and incorporate no intelligence.

LE3D is designed with both research and real-world scenarios in mind. It is an extensible framework and comes with a set of supporting tools for testing and experimentation. Initially, a detection application operates on each ``edge'' IoT node. This application maintains multiple detectors. Each stream received instantiates a detector with multiple estimators, handles the data received, identifies occurred drift, and maintains the voting and adapting windows. An aggregation application collects the results and metadata from the current and neighbouring ``edge'' nodes and decides whether the drift detected is natural or abnormal. The decision is later sent to the ``cloud'' tier for visualisation. All our interactions and messages are exchanged via an MQTT data bus running at the ``edge'' tier and a set of pre-defined topics. A more in-depth explanation of the implementation of LE3D can be found in~\cite{le3d_demo}.

\begin{figure}[t]
    \centering
    \includegraphics[width=1\columnwidth]{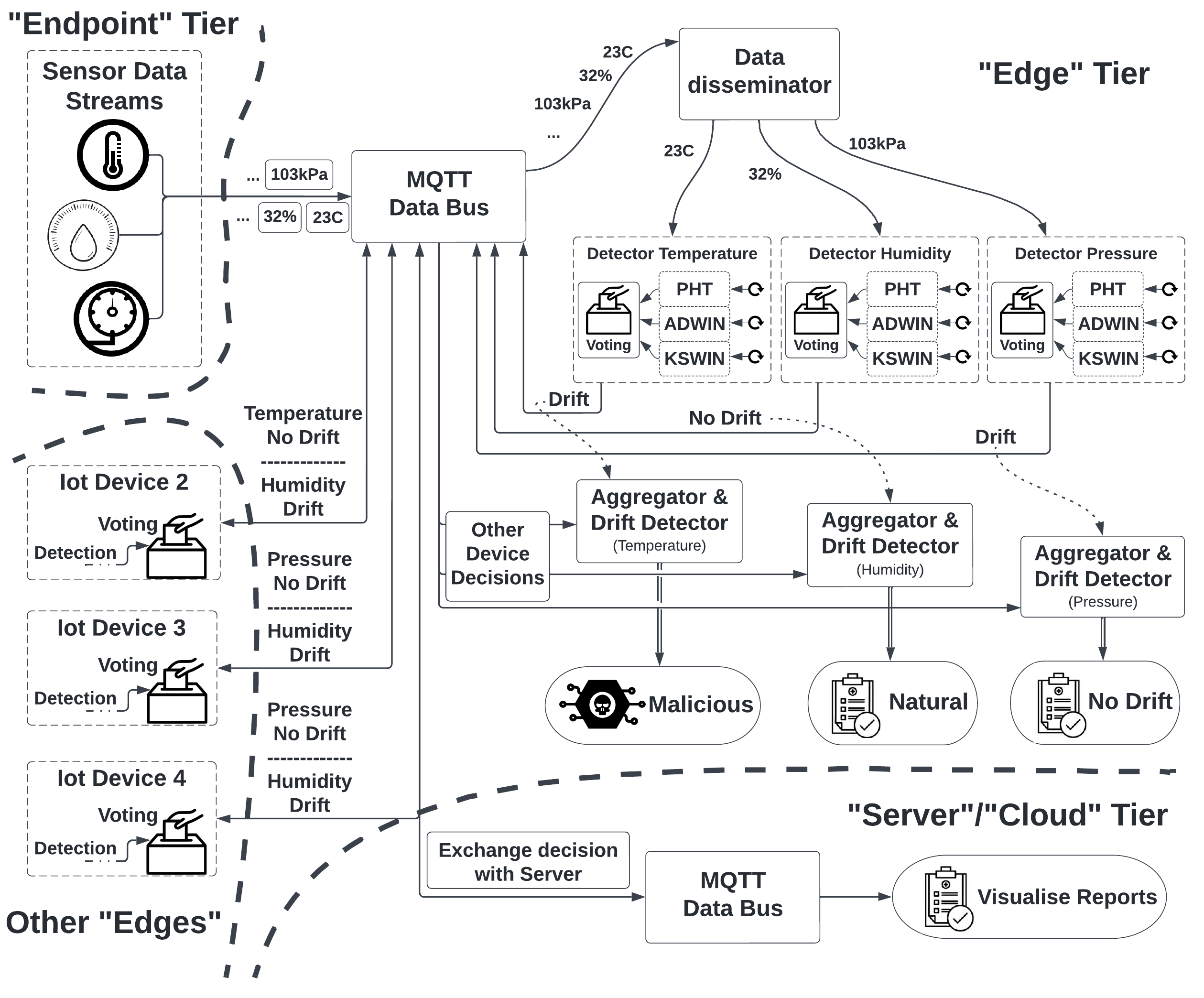}
    \caption{A diagram visualising the different system components and the interactions between them.}
    \label{fig:highleveldiagram}
\end{figure}

As access to real-world drifting endpoints is not always possible, we developed a set of supporting tools for visualisation and experimentation. A streaming application streams ``real-world'' data from a pre-existing dataset (CSV files). An emulator generates ``realistic'' emulated data streams and drifts on demand. A matching application ensures the system's scalability and easier experimentation by matching different detectors with ``streamers'' and ``emulators''. Finally, all emulators expose RESTfull APIs for introducing drift on-the-fly. A high-level representation of the framework and the interactions can be seen in Fig.~\ref{fig:highleveldiagram}.

Our framework was implemented in Python 3.9.12 and is available to the public domain\footnote{https://github.com/toshiba-bril/le3dDataDriftDetector}. The estimators' functionality is based on River online/streaming ML package~\cite{river}. We overrode various functions to adapt them to our framework and achieve the desired functionality introduced earlier (Sec.~\ref{sec:framework}). The linear regression, the grid search and the statistical calculations are based on SciPy and NumPy open-source libraries. The MQTT messaging is based on the Paho MQTT client implementation provided by Eclipse~\cite{paho}. All RESTFul APIs are developed with Flask. The non-linear regression was performed using the ZunZun curve fitting library~\cite{zunzun}. The rest of the framework was developed in-house. Our codebase was built on a Raspberry Pi (RPi) Compute Module 3b+~\cite{RPI_3_COMPUTE}, with a BCM2837B0 Cortex-A53 64-bit \SI{1.2}{\giga\hertz} System-on-a-Chip (SoC) and \SI{1}{\giga\byte} of RAM. This RPi was chosen as a representative resource-constrained IoT device.

\subsection{System Scalability}
Even though evaluating the scalability of our solution is out of the scope of this work, in this section, we address some ideas considered during our implementation phase. First, as the lightweight operation is essential for real-world frameworks, the statistical models and software libraries considered were validated for their resource utilisation before their integration into the system, always considering the accuracy of the predictions as well. Second, in terms of response and execution time, the requests that can be served per second, and memory usage, we quantified the system's performance in Sec.~\ref{subsec:lightweightperformance}. Third, regarding network usage, reducing the exchange of sensor data and exchanging only the voting decisions not only preserves the data integrity and confidentiality but can also reduce the exchange of network data. Fourth, in terms of the horizontal scaling of the system, as discussed in Sec.~\ref{subsec:naturalmalicious}, the voting decisions should be compared and exchanged with close proximity or look-alike neighbours (with common features). 

Even though not considered at this stage, a neighbour discovery mechanism can optimise the exchange of voting decisions even for large-scale deployments. Finally, in terms of the number of sensor streams supported per detector, our multithreaded implementation makes the operating system's kernel and the number of threads supported there the only limiting factor.

\subsection{IoT Endpoints and Sensors}\label{subsec:uob_sec}

\begin{figure}[t]
    \centering
    \includegraphics[width=0.9\columnwidth]{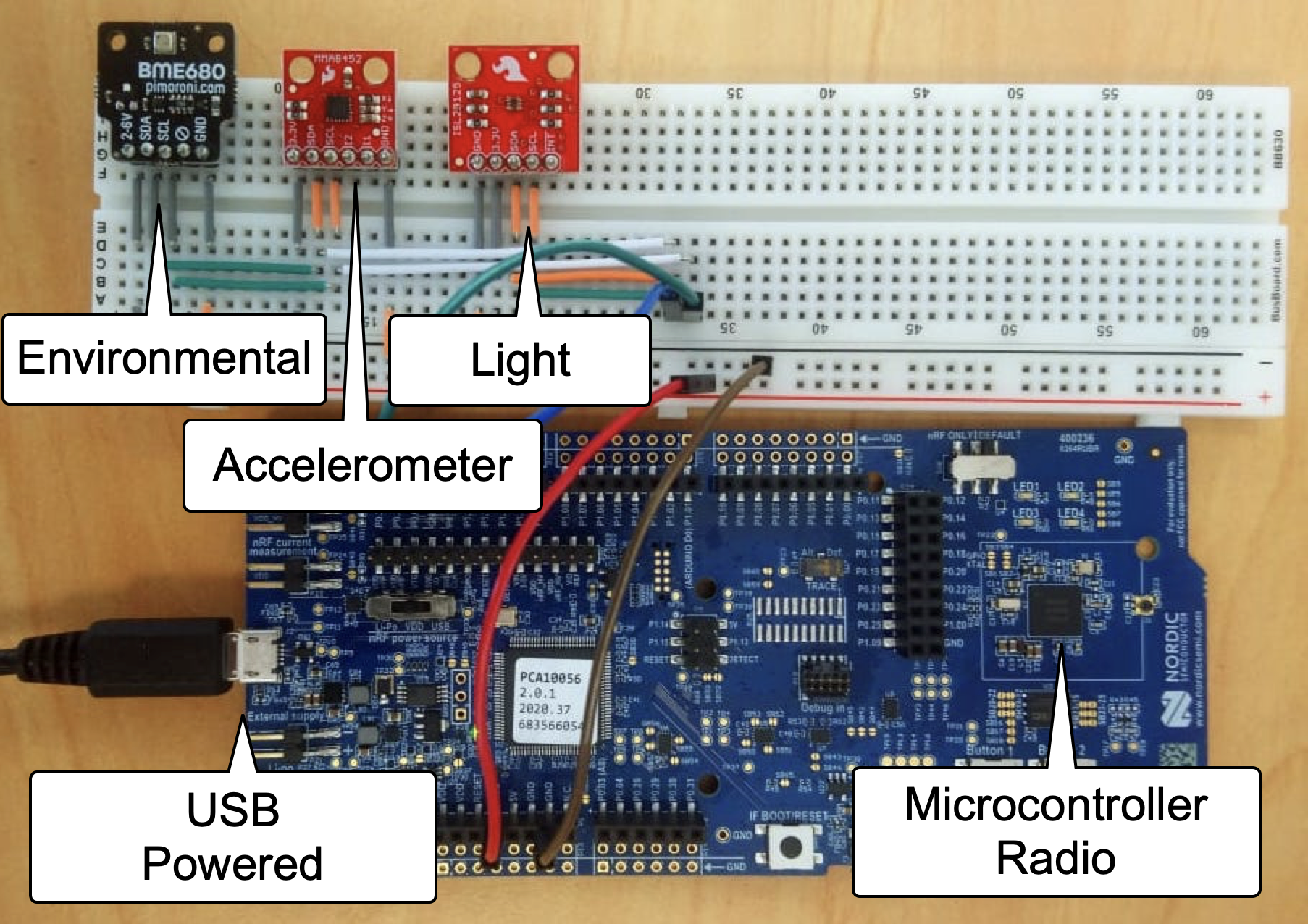}
    \caption{Our developed IoT Endpoint and Sensors boards.}
    \label{fig:sensor_device}
\end{figure}

The data used for the initial model was collected at the Communication Systems \& Networks Laboratory at the University of Bristol between \SIrange[range-phrase=--]{15}{22}{} February 2022. The data collection effort went through an ethics approval process. A total of eight IoT endpoint devices (Fig.~\ref{fig:sensor_device}) were deployed in an office environment, all equipped with commercial off-the-shelf sensors and a wireless microcontroller. The devices were spread around the lab and office space areas, with roughly \SIrange[range-phrase=--]{10}{12}{} researchers usually present during normal working hours. A copy of the dataset can be found at {\tt\small github.com/jpope8/synergia\_datadrift\_dataset}. All devices were USB powered and were equipped with:

\begin{enumerate}
    \item A Nordic nRF52480 Bluetooth SoC.
    \item A Light Sensor (ISL29125): Collects both colour and light intensity values.
    \item An Accelerometer Sensor (MMA8452Q).
    \item An Environmental Sensor (BME680): Collects temperature, humidity, pressure, \& gas (VOC/CO$_2$) values.
\end{enumerate}

The endpoints collected sensor samples every \SI{10}{\second}, the data were stored on a nearby root controller/desktop and were exchanged via an IEEE 802.15.4 Time Synchronised Channel Hopping (TSCH) mesh network. This dataset was used for our real-world performance investigation to generate real-world ``streams'' of data, construct realistic emulators for demonstrating drifting behaviour and for the initialisation of the estimators (as described in Secs.~\ref{subsec:initialisation} and~\ref{subsec:implementation}).

\section{Performance Evaluation}\label{sec:results}
Our performance investigation is two-fold. Initially, we compare LE3D against PWPAE~\cite{PWPAE} and the performance of each individual estimator introduced in LE3D. PWPAE was chosen due to its similarity to our framework, the high accuracy it achieves, and its lightweight nature. Later, we conduct a detailed performance profiling to measure LE3D's execution time and perceived resource utilisation. PWPAE is an ensemble drift detection based on four classifiers, these being the Streaming Random Patches (SRP) classifier (using either ADWIN or Drift Detection Method (DDM) as its base estimators), and Adaptive Random Forest (ARF) classifier (again using either ADWIN or DDM as its base estimators). The four classifiers are, by default, ensemble methods using multiple instances of the same estimator in the background. Following the authors' recommendation, PWPAE classifiers were configured with three instances of each estimator resulting in twelve estimators in total contributing to the final result.

Our evaluation is based on both real-world and emulated data. We conducted experiments for three different sensor types, i.e., temperature, pressure and humidity. The data streams are generated on a desktop PC and fed into the detectors.  We generate a data sequence of $40$ timeslots for each emulated stream. Each timeslot has a random length of $l$ samples with $l \in [500,1500]$. This generates approximately \textasciitilde$40$k samples per experiment. Each timeslot is assigned a type with equal probabilities, i.e., ``normal'', ``incremental'', and ``abrupt''. The first timeslot is always ``normal'', as it corresponds to the time frame that the estimators are initialised. The averaged outcome of $1000$ experiments will be presented later in this section.

At the beginning of each experiment with define $q \in \left \{ \nicefrac{\sigma^2}{3}, \nicefrac{\sigma^2}{2}, \sigma^2, 2 \, \sigma^2, 3 \, \sigma^2, 4 \, \sigma^2, 5 \, \sigma^2  \right \}$ where $\sigma^2$ is the variance of each sensor type (Table~\ref{tab:hyperparameters}). $q$ is fixed for the rest of the experiment. For each timeslot, a value is drawn from a distribution that dictates the drift per timeslot. More specifically, for an abrupt drift, samples are drawn from $x \in \mathcal{N}(\mu^\prime + Q,\,\sigma^{2})$ where $Q \in [-q, q]$. For an incremental drift, we calculate the step $z$ as $z = \nicefrac{Q}{l}$, and the samples are given from the equation $y(x) = z \, x + x$, where $x \in \mathcal{N}(\mu^\prime,\,\sigma^{2})$.

\begin{table}[t] 
\renewcommand{\arraystretch}{1.12}
\centering
    \caption{Hyperparameters for all Estimators and Sensors.}
    \vspace{1mm}
    \begin{tabular}{r||c|c|c||l}
    \textbf{Sensor} & \textbf{ADWIN} & \textbf{PHT} & \textbf{KSWIN} & \textbf{Stats}  \\ \hline \hline
    \multirow{3}{*}{Temperature} & \multirow{3}{*}{$\scriptstyle \delta = 0.44$} & \multirow{2}{*}{$\scriptstyle \beta = 0.095$} & $\scriptstyle \alpha = 0.001$ & \multirow{3}{*}{\parbox{1.3cm}{$\scriptstyle \mu^\prime = 20.32^o$ $\scriptstyle \sigma^2 = 1.178$}} \\
    & & & $\scriptstyle L_{\mathrm{R}} = 300$ & \\
    & & $\scriptstyle\lambda = 480$ & $\scriptstyle L_{\mathrm{\Omega}} = 30$ & \\ \hline
    \multirow{3}{*}{Humidity} & \multirow{3}{*}{$\scriptstyle\delta = 0.44$} & \multirow{2}{*}{$\scriptstyle\beta = 0.095$} & $\scriptstyle\alpha = 0.001$ & \multirow{3}{*}{\parbox{1.3cm}{$\scriptstyle \mu^\prime = 30.14\%$  $\scriptstyle\sigma^2 = 0.966$}} \\
    & &  & $\scriptstyle L_{\mathrm{R}} = 300$ \\
    & & $\scriptstyle \lambda = 560$ & $\scriptstyle L_{\mathrm{\Omega}} = 30$ \\ \hline
    \multirow{3}{*}{Pressure} & \multirow{3}{*}{$\scriptstyle \delta = 0.34$} & \multirow{2}{*}{$\scriptstyle \beta = 2.9$} & $\scriptstyle \alpha = 0.0001$ & \multirow{3}{*}{\parbox{1.5cm}{$\scriptstyle \mu^\prime = 102.4\mathrm{kPa}$ $\scriptstyle \sigma^2 = 224.52$}} \\
    & & & $\scriptstyle L_{\mathrm{R}} = 300$ \\
    & & $\scriptstyle \lambda = 29000$ & $\scriptstyle L_{\mathrm{\Omega}} = 30$ \\
    \end{tabular}
\label{tab:hyperparameters}
\end{table}

\begin{figure}[t]
    \centering
    \includegraphics[width=\columnwidth]{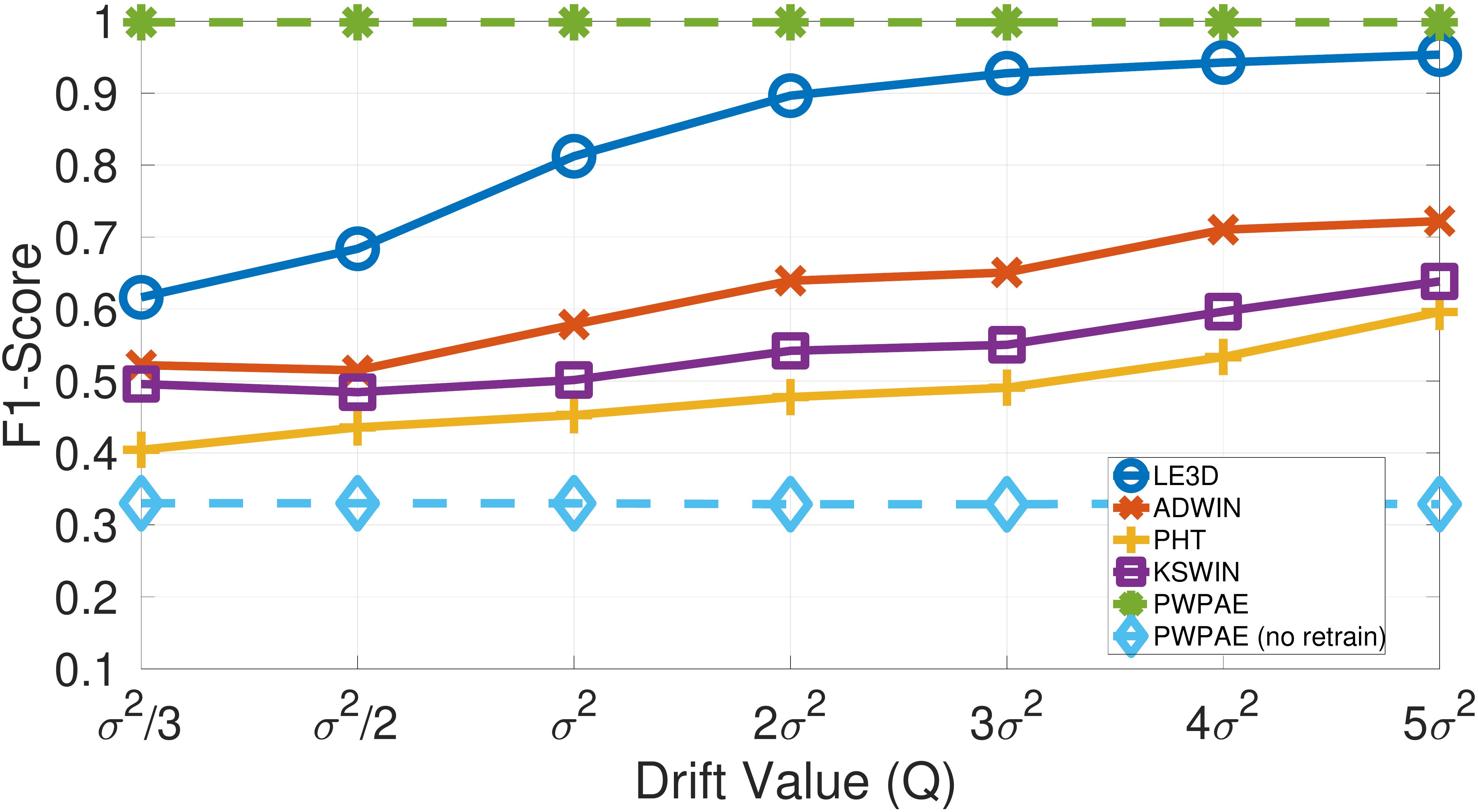}
    \caption{F1-score for humidity and all drift values.}
    \label{fig:humidity}
\end{figure}

\begin{figure}[t]
    \centering
    \includegraphics[width=\columnwidth]{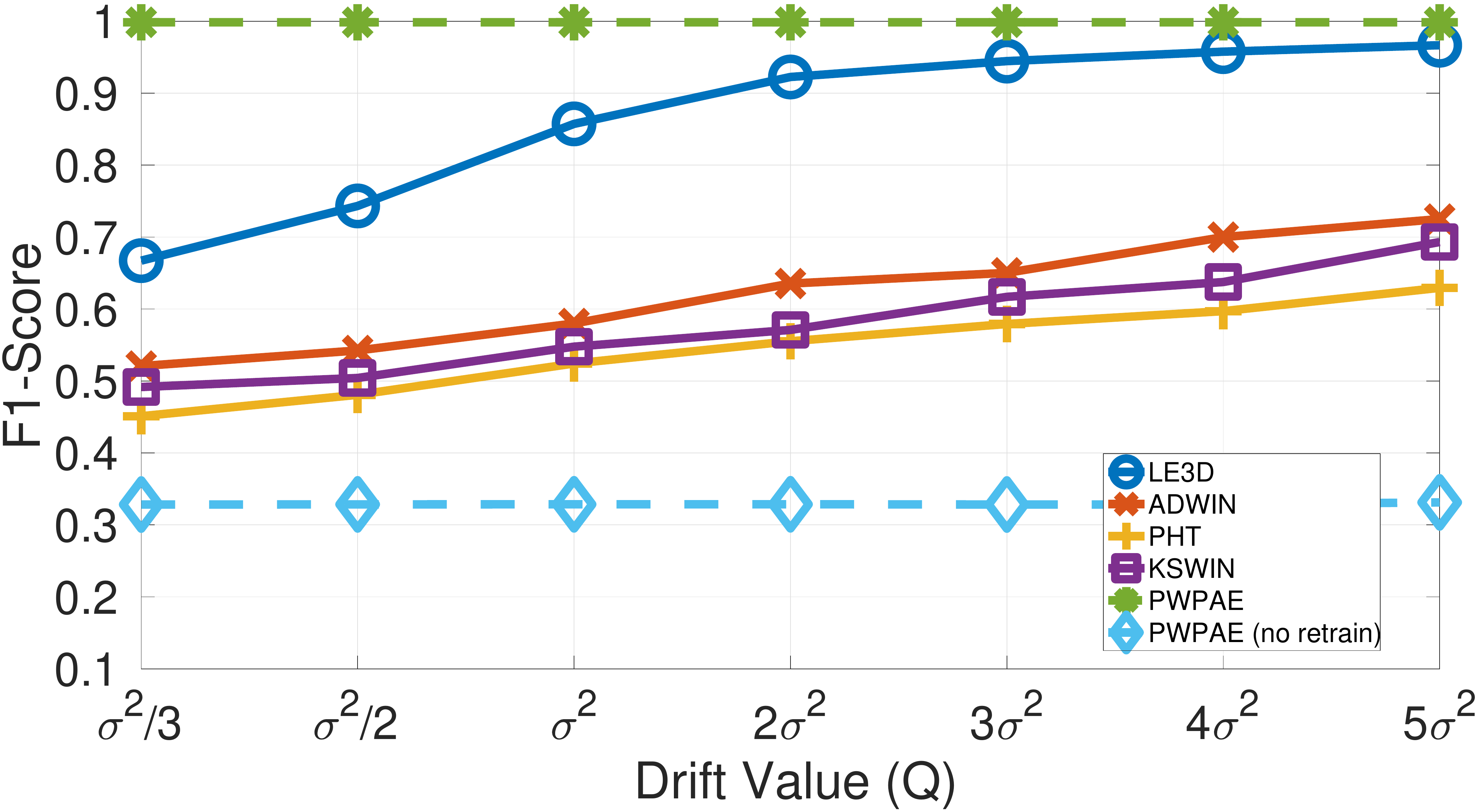}
    \caption{F1-score for temperature and all drift values.}
    \label{fig:temperature}
\end{figure}

\begin{figure}[t]
    \centering
    \includegraphics[width=\columnwidth]{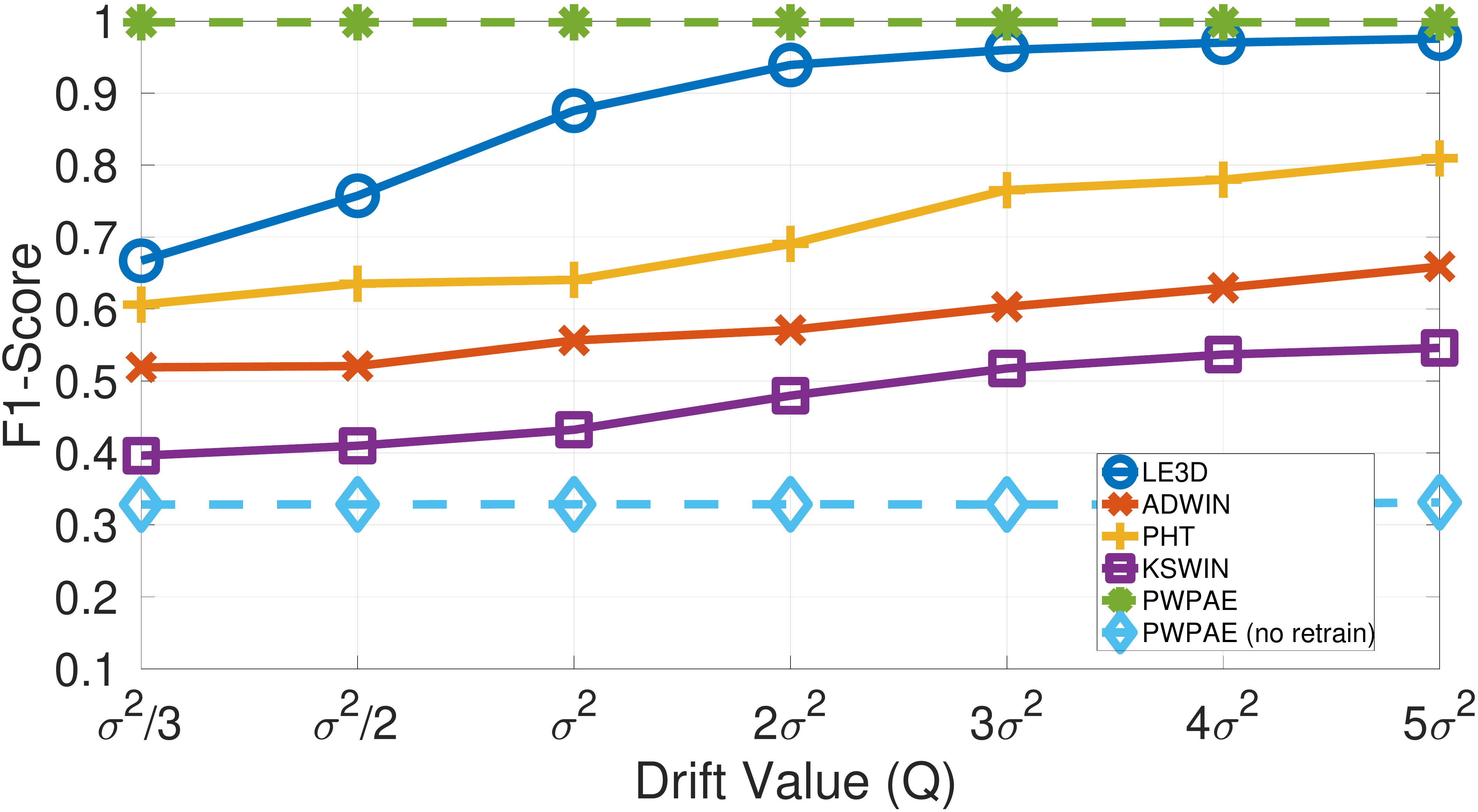}
    \caption{F1-score for pressure and all drift values.}
    \label{fig:pressure}
\end{figure}

     

\subsection{Experimental Evaluation}
As discussed, LE3D execution starts with initialising the estimators' hyperparameters. In Table~\ref{tab:hyperparameters}, we present the grid search optimised hyperparameters (as described in Sec.~\ref{subsec:initialisation}), and the $\mu^\prime$ and $\sigma^2$ for all sensors. Investigating the PWPAE codebase, we pinpointed that after the initial training, the authors also continued the training during the inference phase. Their model is updated on a per-sample basis using the expected labels and not the predicted ones (as is usually the case in supervised learning methods). Such an implementation is rather infeasible for a real-world system, as the expected labels will never be available for online training. Thus for our evaluation, we assessed PWPAE in two different scenarios: 1) with the online training enabled, 2) with the online training disabled.

Our results are presented in Figs.~\ref{fig:humidity}~-~\ref{fig:pressure}. F1-score was chosen as a good metric due to the imbalanced nature of our data (i.e., more drifts are generated than normal data). Our evaluation show that LE3D performs very well for all sensor types, can provide a generalisable approach, and achieves high F1-scores up to $97\%$. Even for smaller drifts (when $q \geq \sigma^2$), the F1-score slightly decreases but is always above $83\%$.

Considering PWPAE, we see that when the training is enabled, the model always performs great, achieving an F1-score of $99.8\%$. PWPAE requires a very small number of labelled samples to ``learn'' that a drift is introduced within a timeslot. These results, even though on paper, are great, as discussed above, are infeasible to be achieved in a real-world system where the expected labels for each sample are unavailable. This is apparent when the online training is disabled, and PWPAE's F1-score is reduced to \textasciitilde$32\%$. During that scenario PWPAE always reports a ``normal'' behaviour. Investigating further the classifiers proposed from PWPAE, it was identified that they benefit from feature-rich datasets and do not perform well with time-series raw sensor data (like in our case). Of course, even though post-processing the data is feasible, it will introduce more overhead in the system. 

Finally, considering the individual estimators, we see that PHT performs better for pressure streams. This is because the observed $\mu_{\mathbb{P}}$ is a very prominent component in the PHT algorithm. When the absolute value of a sensor steams is not modified significantly (e.g., temperature and humidity), KSWIN and ADWIN outperform PHT. However, as observed, all estimators independently perform worse than our ensemble framework. From the results, it is evident that the collective decision taken using our voting mechanism can enhance the detection accuracy. From the above, and considering the requirements for real-world deployments, LE3D significantly outperforms individual estimators and other state-of-the-art solutions when only raw data are available.




\subsection{Execution Time and Resource Consumption Evaluation}\label{subsec:lightweightperformance}
We profiled LE3D and PWPAE codes, measuring the execution time (Table~\ref{tab:performance_comparison}) and the resource utilisation (Table~\ref{tab:performance}). Both experiments were conducted on an RPi 3b+ with a BCM2837B0 Cortex-A53 64-bit \SI{1.2}{\giga\hertz} SoC and \SI{1}{\giga\byte} of RAM. As discussed in Sec.~\ref{subsec:implementation}, the edge nodes host an MQTT broker and the detection frameworks. 

\begin{table}[t] 
\renewcommand{\arraystretch}{1.12}
\centering
    \caption{Execution Time Comparison against PWPAE~\cite{PWPAE}.}
    \vspace{1mm}
    \begin{tabular}{R{2.5cm}|M{1.6cm}|M{1.8cm}}
    \textbf{Test} & \textbf{Detector Initialisation} & \textbf{Sample processing}  \\ \hline \hline
    PWPAE & \multirow{2}{*}{\textasciitilde\SI{4.5}{\milli\second}} & \SIrange[range-phrase=--,range-units=single]{6500}{8500}{\micro\second} \\ \cline{1-1}\cline{3-3}
    PWPAE (no training) &  & \SIrange[range-phrase=--,range-units=single]{2000}{2800}{\micro\second} \\ \hline
    Our approach & \textasciitilde\SI{3}{\milli\second} & \SIrange[range-phrase=--,range-units=single]{150}{600}{\micro\second} \\ \hline
    \end{tabular}
\label{tab:performance_comparison}
\end{table}

\begin{table}[t] 
\renewcommand{\arraystretch}{1.12}
\centering
    \caption{Performance Profiling on a Raspberry Pi 3B+.}
    \vspace{1mm}
    \begin{tabular}{R{1.6cm}|M{1.6cm}|M{0.7cm}|M{0.7cm}|M{0.7cm}}
    \textbf{Test} & \textbf{Incoming Sample Rate} & \textbf{CPU} & \textbf{RAM} & \textbf{System Load}  \\ \hline \hline
    Idle RPi & - & $0.1\%$ & $1.4\%$ & 0.0025 \\ \hline
    Just MQTT & - & $0.3\%$ & $1.4\%$ & 0.005 \\ \hline
    Ensemble f/w \& MQTT & - & $0.4\%$ & $7.8\%$ & 0.0075 \\ \hline \hline
    \multirow{7}{1.2cm}{\raggedleft Sampling Rate} & \SI{1}{\hertz} & $0.45\%$ & $7.8\%$ & 0.0075 \\ \cline{2-5}
    & \SI{10}{\hertz} & $0.7\%$ & $7.8\%$ & 0.01 \\ \cline{2-5}
    & \SI{100}{\hertz} & $1.2\%$ & $7.9\%$ & 0.015 \\ \cline{2-5}
    & \SI{500}{\hertz} & $3.9\%$ & $8.1\%$ & 0.035 \\ \cline{2-5}
    & \SI{1000}{\hertz} & $7.6\%$ & $8.7\%$ & 0.08 \\ \cline{2-5}
    & \SI{2000}{\hertz} & $9.85\%$ & $9.8\%$ & 0.12 \\ \cline{2-5}
    & \SI{4000}{\hertz} & $19.5\%$ & $11.4\%$ & 0.24 \\ \hline
    \end{tabular}
\label{tab:performance}
\end{table}

As execution time, we quantify: 1) the time required for initialising a detector (e.g., when a new sensor stream is fed into the frameworks), and 2) the time required to process an individual sample. As resource utilisation, we measured the CPU and RAM utilisation (using \textsl{ps} command on Unix, sampled every \SI{0.1}{\second}), and the average system load (using \textsl{uptime} at the end of an experiment). The values are normalised to the four-core architecture of the RPi 3b+ (i.e., divided by four). Each experiment lasted for \SI{20}{\min} capturing the code profile between the \SIrange[range-units=single]{5}{20}{\min} range. 

Table~\ref{tab:performance_comparison} summarises the execution time in comparison with PWPAE. PWPAE was evaluated for both scenarios (with and without online training). When compared we see that LE3D requires \textasciitilde\SI{3}{\milli\second} for initialising a detector, whereas PWPAE requires \textasciitilde\SI{4.5}{\milli\second}. This is not a big difference considering that a detector is initialised only once when a new sensor stream is received. However, the individual sample processing time presents a significant difference. Our approach requires about 20 times less time (comparing the average values) when compared to PWPAE with online training and less than eight times when the training is disabled. As seen, the performance of PWPAE degrades significantly when the training is disabled. Considering the case where the training is enabled, it can be approximated that PWPAE can process \textasciitilde$150$ $\nicefrac{\mathrm{samples}}{\mathrm{s}}$. 

LE3D is later evaluated with up to $4000$ $\nicefrac{\mathrm{samples}}{\mathrm{s}}$ (Table~\ref{tab:performance}), a value significantly greater than PWPAE. As seen, our framework is lightweight and achieves low CPU, i.e., $<10\%$ and $<20\%$ for \SI{2000}{\hertz} and \SI{4000}{\hertz} sampling rates, and RAM utilisation, i.e., always less than $<11.5\%$, even when the number of samples increases. In our framework, historic sensor samples are kept after processing, increasing the RAM usage slightly. This can be regulated by discarding historical data after expiration or when a pre-defined queue is full. Even though the system is still not saturated, we can see that at \textasciitilde$2000$ $\nicefrac{\mathrm{samples}}{\mathrm{s}}$, we approach the limits of our implementation, if a real-time operation is required. Investigating further, we concluded that the single-threaded implementation of the Eclipse Paho MQTT library in Python creates a bottleneck on the number of samples one can subscribe to per second. Workarounds will be considered as future extensions of the system. This could either be replacing the Paho MQTT library with another implementation, using a different messaging protocol, or subscribing to batches of data, keeping them in a separate queue, and processing them asynchronously from when they arrive.


\section{Conclusion}\label{sec:conclusion}
This paper presents LE3D; a lightweight ensemble data drift detection framework for resource-confined IoT environments. Working in a distributed two-tier hierarchical fashion can detect irregularities in the received sensor streams and classify them as natural or abnormal. Our framework is generalisable and performs with high accuracy for different sensor types and data streams, achieving up to 97\% in F1-score. Its strength relies on the system dynamically adapting to new sensor streams without any accuracy reductions, while the collective decisions taken from different devices can provide more in-depth knowledge on the types of drift observed. Moreover, its distributed nature and the fact that the data never leave the device preserves the data confidentiality and integrity. Compared to other state-of-the-art solutions, we can see that the detection accuracy is almost on par with more resource-heavy methods. Furthermore, by conducting an extensive performance profiling of our proposed framework, we demonstrated its lightweight operation in a resource-constrained IoT device. The detection accuracy, the minimal overhead introduced from the implementation, and the framework's scalability make it a great candidate for data drift detection frameworks for real-world IoT sensor applications.

\section*{Acknowledgment}
This work was supported in part by Toshiba Europe Ltd. and in part by the SYNERGIA project (grant no. 53707, UK Research and Innovation, Innovate UK).

\bibliographystyle{IEEEtran}
\bibliography{IEEEabrv,bib.bib}

\end{document}